\pdfoutput=1

\documentclass[11pt]{article}

\usepackage[final]{emnlp2021}

\usepackage{times}
\usepackage{latexsym}

\usepackage[T1]{fontenc}

\usepackage[utf8]{inputenc}

\usepackage{microtype}

\title{DiS-ReX: A Multilingual Dataset for Distantly Supervised Relation Extraction}
\author{Abhyuday Bhartiya \footnotemark \\ 
  Indian Institute of Technology \\
  New Delhi, India \\
  \small \texttt{bhartiyabhyuday@gmail.com} \\\And
 Kartikeya Badola \footnotemark[\value{footnote}] \\
Indian Institute of Technology \\ 
  New Delhi, India \\
  \small \texttt{kartikeya.badola@gmail.com} \\ \And
  Mausam \\
Indian Institute of Technology \\ 
  New Delhi, India \\
   \small \texttt{mausam@cse.iitd.ac.in} \\ }

\begin{document}
\maketitle

\begin{abstract}
Distant supervision (DS) is a well established technique for creating large-scale datasets for relation extraction (RE) without using human annotations. However, research in DS-RE has been mostly limited to the English language. Constraining RE to a single language inhibits utilization of large amounts of data in other languages which could allow extraction of more diverse facts. Very recently, a dataset for multilingual DS-RE has been released. However, our analysis reveals that the proposed dataset exhibits unrealistic characteristics such as 1) lack of sentences that do not express any relation, and 2) all sentences for a given entity pair expressing exactly one relation. We show that these characteristics lead to a gross overestimation of the model performance. In response, we propose a new dataset, DiS-ReX, which alleviates these issues.  Our dataset has more than 1.5 million sentences, spanning across 4 languages with 36 relation classes + 1 no relation (NA) class. We also modify the widely used bag attention models by encoding sentences using mBERT and provide the first benchmark results on multilingual DS-RE. Unlike the competing dataset, we show that our dataset is challenging and leaves enough room for future research to take place in this field.

\end{abstract}
{\let\thefootnote\relax\footnotetext{{* Equal Contribution}}}

\section{Introduction}




Relation extraction (RE) is an important subtask of information extraction. The goal is to identify the relation $R$ between a pair of entities $(e1,e2)$ given context $C$, where $C$ is some text mentioning the two entities. Creating RE datasets using human annotation can be cumbersome, as a result of which most fully supervised datasets are small in size. \citet{mintz2009distant} proposed creating relation extraction datasets by using distant supervision (DS-RE). DS-RE is bag-level classification task where a bag of an entity pair $(e1,e2)$ is defined as the set of all sentences in the dataset which mention both $e1$ and $e2$. If $e1$ and $e2$ have a relation $r$ according to a knowledge base (KB), then the entire bag of $(e1,e2)$ is associated with the label $r$.



Research in DS-RE has been mostly limited to the English language due to unavailability of large multilingual datasets. Since same facts about the world can be represented in different languages, multilingual training of RE models can have several benefits. Having a single model for multilingual tasks vis-a-vis one model for each language 1) allows for cross-lingual knowledge transfer which improves performance across all languages \citep{zoph-etal-2016-transfer,feng2020language}, and 2) is a more efficient method of capturing consistent semantics across languages \cite{lin2017neural}



RELX-Distant \cite{koksal2020relx} is the first multilingual dataset for distantly supervised relation extraction with sentences in 5 languages. Our analysis reveals some critical flaws in this dataset, which make it unsuitable as a reliable DS-RE benchmark:
\begin{enumerate}
    \item It has no negative samples i.e sentences without any possible relationship between a given entity pair
    \item Relation classes are semantically far apart. There exists no entity pair that has more than one possible label in the relation set, even under the distant supervision scheme.
    \item The dataset is extremely imbalanced. More than 50\% bags belong to \textit{country} relation type.
\end{enumerate}

These unrealistic characteristics grossly overestimate an RE model's performance. Our preliminary analysis with a simple mBERT based model achieves an AUC of 0.98 and Micro F1 of 0.95 on the test set after only 5 epochs of training. These issues make the benchmark unsuitable for spurring further research in the field of multilingual DS-RE.

In response, we contribute a more realistic benchmark dataset for the task called DiS-ReX. Our contributions are as follows:
\begin{enumerate}
    \item DiS-ReX is a multilingual distantly supervised relation extraction dataset over 4 languages: English, German, Spanish, French. Our dataset has more than 1.5 million instances. Each instance can be one of 37 relation classes (36 positive and 1 No Relation class)
    \item We also provide the first benchmark numbers on multilingual distantly supervised RE using a combination of mBERT \cite{devlin-etal-2019-bert} encoding with Intra-Bag Attention \cite{lin2016neural} 
\end{enumerate}
Our dataset has similar percentage of no relation (NA) examples as in NYT dataset, which is a standard dataset for English DS-RE. The imbalance among our positive classes is much less compared to RELX distant (almost an order of magnitude improvement). Using our baseline model, we achieve an AUC of 0.8 and Micro F1 of 73\% on the test set. This suggests that unlike the RELX dataset, our dataset is not trivial to optimize on and has the potential to act as a useful and challenging dataset for the task. We publicly release our dataset and baseline model\footnote{The dataset and code for extracting it can be found at \url{https://github.com/dair-iitd/DiS-ReX}}.



\section{Previous Works}

For multilingual supervised relation extraction, ACE05 dataset \cite{ace2005} and KLUE dataset \cite{klue2010} have been used as standard benchmarks. Both of these datasets have less number of examples(\textasciitilde10k). \citet{mintz2009distant} proposed distant supervision as an alternative to large-scale human annotation. To deal with the wrong labelling problem, \citet{riedel2010modeling} apply the "at-least one" assumption and \citet{hoffmann2011knowledge} and \citet{surdeanu2012multi} remodelled distantly supervised RE as a multi-instance multi-label (MI-ML) classification task. Under the MI-ML framework, the goal of the extractor is to predict all possible relations from the set of relation classes for a given bag which contains multiple sentences for the same entity pair. \citet{riedel2010modeling} also introduce the New York Times (NYT) dataset built using distant supervision, which serves as an important benchmark for English DS-RE. RELX-Distant\cite{koksal2020relx} is the first dataset for multi-lingual DS-RE but it suffers from several drawbacks discussed in previous section. Moreover, the authors did not publish any benchmark numbers on the RELX-Distant dataset, and instead used distant supervision for pre-training a downstream supervised RE system.

One of the first deep neural networks for DS-RE are piece-wise CNN (PCNN) based methods \cite{zeng2015distant}. \citet{lin2016neural} combine PCNN with intra-bag attention in which a trainable relation embedding attends over all sentences in a bag and generates a bag-level representation which is used for prediction.

\citet{lin2017neural} and \citet{wang2018adversarial} proposed extension of bag-attention models for bilingual datasets. However, adoption of these models to multiple languages has been limited due to lack of multi-lingual DS-RE datasets Instead of using separate sentence encoders for each language, we modify the bag attention model by encoding sentence using a common mBERT model. This serves as a baseline benchmark for our dataset.



\section{Methodology}

\subsection{Dataset creation pipeline}
We first harvest a large number of sentences in English, French, Spanish and German using Wikipedia. We then hypothesize relations using distant supervision by aligning sentences with DBpedia KB \cite{lehmann2015dbpedia} which is a large-scale multilingual KB extracted from Wikipedia. For dataset creation, we build a general pipeline that can be extended to obtain datasets from other text corpora besides Wikipedia. The specific steps are as follows: 

\begin{enumerate}
  \item Wikimedia dumps for each language are downloaded and split into sentences. Entities present in the sentences are detected using a language-specific NER tagger \cite{honnibal2017spacy}.
  \item We use different DBpedia language editions for sentences from different languages. This gives us a better and increased coverage on entities that are local to different language speaking parts of the world. 
  \item We fuse the KBs of different language editions by finding the Wikidata ID for each entity. Wikidata IDs are consistent across languages and allows us to establish equivalence between entities like \textit{USA} and \textit{Estados Unidos de América}.
  \item Entities detected in sentences are aligned with the fused KB by string matching. We only select entities for which we obtain an exact match in the KB.
  \item For each entity pair in a sentence, we search for a relation between them in the knowledge base. If a relation is found, that instance is labelled with it, otherwise label is "NA". 
\end{enumerate}
%

 We then select the top 50 positive relations classes based on number of bags from all languages combined. Relation types which do not have more than 50 bags in each of the 4 languages are discarded. We end up with 36 positive relation classes. We then add the bags of entity pairs which have no relation between them. We filter bags with "NA" label to achieve similar percentage of instances as in the NYT Dataset (around 70\%)
 
 To make the dataset more balanced, we limit the number of bags for each relation type in each language to a max of 10,000. This helps curb the skew due to relation types such as $country$ and $birthPlace$. During the filtering process, we ensure that bags of entity pairs common across more than 1 language are not removed so that we have an abundant number of cross-lingual bags. Models can take advantage of such bags for establishing representation consistency across languages \cite{wang2018adversarial}.

Key statistics of our dataset are shown in Table \ref{tab:1}. Positive relations classes in our dataset are :

\textit{artist, associatedBand, author, bandMember, birthPlace, capital, city, country, deathPlace, department, director, formerBandMember, headquarter, hometown, influenced, influencedBy, isPartOf, largestCity, leaderName, locatedInArea, location, locationCountry, nationality, predecessor, previousWork, producer, province, recordLabel, region, related, riverMouth, starring, state, subsequentWork, successor, team}

For evaluation on this dataset,  we combine sentences from all languages and create two types of splits. We call one as \textbf{unseen} split and other as \textbf{translation} split. For unseen split, bags in the test set and training set are mutually exclusive. For translation split, mutual exclusion only holds between train and test bags of the same language but there can be common bags between train bags of one language and test bags of other. Our train, validation and test sets are in the ratio $70:10:20$

Translation and Unseen splits measure different capabilities of an extractor. Unseen split measures how well an extractor is able to generalize to new entity pairs whereas translation split measures how well an extractor is able to memorize and recall facts learnt through one language, when tested on a different language.

\subsection{BERT + Bag-Attention Baseline}

We now describe our baseline Multilingual DS-RE model. Let $B = \{\beta_1,\beta_2 ... \beta_l\}$ denote bag of sentences in $l$ different languages with same entity pair $(e_1,e_2)$ and label $r$. Here $\beta_i = \{x^1_i,x^2_i .. x^{n_i}_i\}$ is set of sentences in language $i$ with entity pair $(e_1,e_2)$ and label $r$ and contains $n^i$ sentences. Using our model, we obtain probabilities $p(r|B,\theta)$ which measures likelihood of $r$ being a label for bag B.

\subsubsection{BERT Encoder}
To obtain a distributed representation of a sentence $x$, we use mBERT. In order to encode positional information into the model we use Entity Markers scheme introduced by \cite{soares2019matching}. We add special tokens $[E1]$ , $[\backslash E1]$ to mark start and end of the head entity and $[E2]$ , $[\backslash E2]$ to mark start and end of the tail entity. This modified sentence is fed into a pretrained BERT model and the output head and tail tokens are concatenated to get the final sentence representation $\tilde{\textbf{x}}^j_i$ for each sentence $x^j_i$ in our bag. 

\begin{table}
\centering
\begin{tabular}{lcccc}
\hline
\textbf{Language} & \textbf{\# sentences} & \textbf{\#  non-NA bags} \\
\hline
\verb|English| & {532499} & {66932} \\
\verb|French| & {409087} & {83951} \\
\verb|Spanish| & {456418} & {80706} \\ 
\verb|German| & {438315} & {45908} \\ 
\hline

\end{tabular}
\caption{Key statistics for DiS-ReX}
\label{tab:1}
\end{table}

\subsubsection{Intra Bag Attention}
To obtain representation of bag $B$, we apply selective sentence-level attention \cite{lin2016neural}. We obtain real-valued vector $\tilde{\textbf{B}}$ for the bag as a weighted sum of sentence representations $\tilde{\textbf{x}}^j_i$ :\newline
\begin{center}
$\tilde{\textbf{B}}$ = $\sum_{i,j}\alpha^j_i*\tilde{\textbf{x}}^j_i$ 
\end{center}
where $\alpha^j_i$ measures attention score of $\tilde{\textbf{x}}^j_i$ with a specific relation $\textbf{r}$ :-
\begin{center}
$\alpha^j_i$ = $\frac{exp(\tilde{\textbf{x}}^j_i \cdot \textbf{r})}{\sum_{k,l} exp(\tilde{\textbf{x}}^k_l \cdot \textbf{r})}$ 
\end{center}
This reduces the effect of noisy labels on the final bag representation.

Finally, we obtain conditional probability $p(r | B,\theta) = softmax(\textbf{o})$. Here we obtain $\textbf{o}$ which represents scores for all relation types.

\begin{center}
$\textbf{o} = \textbf{R}\tilde{\textbf{B}} + \textbf{d}$ 
\end{center}

\textbf{R} is the matrix of relation representations. Our objective function is the cross-entropy loss and is defined as follows :-
\begin{center}
$L(\theta) = \sum_{i = 1}^{b} p(r_i | B_i,\theta) $ 
\end{center}

where $b$ denotes the number of bags in our training data
\section{Experiments and Analysis}

\begin{table*}
\centering
\begin{tabular}{|*{5}{c|} }
    \hline
Language    & \multicolumn{2}{c|}{DiS-ReX} & \multicolumn{2}{c|}{RELX-Distant}  \\
\hline
   & Max \% & Efficiency  &   Max \% & Efficiency  \\
\hline
English   &   14.98  &   0.8748  &   50.28  &  0.5342 \\
    \hline
French   &   11.46 &   0.8713  &   55.49  &  0.4763  \\
    \hline
Spanish   &  12.15  &   0.8786   &   52.05    & 0.5551           \\
\hline
German   &  17.27     &  0.7995 & 46.55   &   0.5237     \\
    \hline
\end{tabular}
\caption{Comparing bag-wise imbalance in RELX-Distant and DiS-ReX}
\label{tab:2}
\end{table*}


\subsection{Comparison of datasets}
DS-RE is modelled as an MI-ML task. RELX-Distant contradicts the multi-label assumption as there exists no entity-pair which have more than one relation label as the ground truth. This is because relation types in RELX-Distant are not fine-grained and many of them are mutually exclusive. For instance, person-person relations in RELX-Distant are: \textit{mother, spouse, father, sibling, partner}. We see that for a given person-person entity pair, there will almost always be exactly one possible relation in the knowledge base. This is infact the case for all relation classes in the RELX-Distant dataset. Evaluating a classifier on such a dataset is not indicative of how it will perform in the real world setting where the relation types are much more fine-grained. 




Ideally, a dataset for DS-RE should have sufficient number of multi-label bags. Further, instances in such datasets should be evenly distributed between different relation classes so that a model cannot choose to ignore classes with few examples in order to increase its accuracy. Our DiS-ReX dataset has the following attributes:
\begin{itemize}
    \item DiS-ReX has 21642(\textasciitilde 10\%) bags which have more than one relation label. An entity pair can have up to 5 possible relations. An example of a bag with 4 relations:
    \begin{center}
        \textit{('Isaac Newton', 'England'): 
 'http://dbpedia.org/ontology/birthPlace',
  'http://dbpedia.org/ontology/country',
  'http://dbpedia.org/ontology/deathPlace',
  'http://dbpedia.org/ontology/nationality'}
    \end{center}

    \item Moreover, DiS-ReX also has inverse relations (unlike RELX-Distant) which ensures that the model learns that an entity pair should be ordered. Some of the examples are: 
    \begin{center}
        \textit{(successor,predecessor),
        (influenced by, influenced),
        (previous work, subsequent work),
        (associated band, band member)}
    \end{center}
    \item In real world datasets, the model must also learn to predict whether an two entities are even related to each other. Hence, our dataset contains instances of "NA" class with a similar percentage as the NYT dataset. 
    
\end{itemize}


In order to compare the imbalance among non-NA relation classes in DiS-ReX and RELX-Distant, we calculate normalized entropy \cite{shannon1948mathematical}, also known an efficiency, over the distribution of relation classes . For $k$ classes, where $i^{th}$ class has $n_i$ number of instances and the total number of instances across all $k$ classes is $n$ :
\begin{center}
    $Efficiency = - \sum_{i=1}^{k} \frac{\frac{n_i}{n}\log \frac{n_i}{n}}{\log k}$
\end{center}

Efficiency lies between 0 and 1. Higher efficiency means that the class-distribution is closer to a uniform distribution. We report the Efficiency and percentage of instances in largest relation class among non NA relation classes for RELX-Distant and DiS-ReX in table \ref{tab:2}

We find that there is high imbalance in the RELX-Distant dataset. This contributes to how easily a baseline model can obtain close to 0.95 micro-F1 score.

\begin{table*}
\centering
\begin{tabular}{|*{9}{c|} }
    \hline
Language    & \multicolumn{4}{c|}{RELX-Distant} & \multicolumn{4}{c|}{DiS-ReX}  \\
\hline
 & \multicolumn{2}{c|}{Translation} & \multicolumn{2}{c|}{Unseen} &  \multicolumn{2}{c|}{Translation} & \multicolumn{2}{c|}{Unseen}  \\
\hline
   & AUC & Micro F1 &  AUC &   Micro F1 & AUC & Micro F1 &  AUC &   Micro F1 \\
\hline
English   &  0.986     &  0.948 & 0.985  &   0.947   &  0.784   &  0.719 & 0.781     &   0.713       \\
\hline
French   &  0.989     &  0.958 & 0.988  &   0.955   &  0.819   &   0.750 & 0.814    & 0.746           \\
\hline
Spanish   &  0.984     &  0.945 & 0.980  &  0.941  &   0.803   &  0.739 & 0.799    & 0.729           \\
\hline
German   &  0.986     &  0.949 & 0.986  &   0.949   & 0.771  &   0.728 &  0.757   & 0.716          \\
\hline
All languages   &   0.986 &   0.949  & 0.986  &   0.948 & 0.799  &  0.734 &  0.806  &  0.726 \\
    
\hline
\end{tabular}
\caption{Comparing AUC and Micro F1 on test set in RELX-Distant and DiS-ReX}
\label{tab:3}
\end{table*}

\subsection{BERT Encoder+attention baselines}
We run our baseline model of mBERT+bag attention on both DiS-ReX and RELX-Distant. We report AUC and Micro F1 in table \ref{tab:3}


For training we use AdamW optimizer \cite{kingma2017adam, loshchilov2019decoupled}, with lr=0.001, betas=(0.9, 0.999), eps=1e-08. Weight decay is 0.01 for all parameters except bias and layer norm parameters. We follow the training pipeline from \citet{lin2016neural} and set bag size to be 2. This means that in one forward pass, our network will process two sentences together belonging to the same bag. We train our model for 5 epochs on both the splits for both datasets, jointly on all languages. Correct prediction of NA class is not counted in the calculation of Micro F1 and AUC. 

As can be seen in table \ref{tab:3}, our baseline model achieves very high micro-F1 scores on the test set of RELX-Distant. Unlike RELX-Distant, DiS-ReX has lower numbers (and similar to state of the art in monolingual DS-RE), suggesting that DiS-REX is a more realistic and challenging dataset for our task.



\section{Conclusion}
In this paper, we propose DiS-ReX, a novel dataset for DS-RE in 4 languages. We show that it is a more reliable benchmark compared to existing multilingual DS-RE datasets. We also publish first baseline numbers on closed domain multilingual DS-RE. Our dataset has fairly even distribution of classes, includes instances with no-relation between entity-pair and the relation-types selected are fine-grained. We show that these attributes make our dataset more challenging and motivates future research in mulitilingual DS-RE.



\bibliography{emnlp2021}
\bibliographystyle{acl_natbib}
\onecolumn
\appendix

\section{Appendix}
\label{sec:appendix}
\subsection{Cross-Lingual Bags in Dis-ReX}
We designed DiS-ReX to have a large number of cross-lingual bags. We show number of bags common across 2, 3 and all languages in table \ref{tab:5}

\subsection{Qualitative Analysis}
In this section, we give some examples of randomly selected non NA instances in our dataset:
\\
\textbf{English:}
\begin{itemize}
    \item \begin{center}
    \textit{\textbf{Sentence:} another dialect spoken in tioman island is a distinct malay variant and most closely related to riau archipelago malay subdialect spoken in natuna and anambas islands in the south china sea together forming a dialect continuum between the bornean malay with the mainland malay}
    \\
    \textit{\textbf{Entities:} (tioman island, the south china sea)
    \\
    \textbf{Relations:} http://dbpedia.org/ontology/location}
    \end{center}
    \item \begin{center}
    \textit{\textbf{Sentence:} in 2017 jenny durkan was elected as the first openly lesbian mayor of seattle}
    \\
    \textit{\textbf{Entities:} (jenny durkan, seattle)
    \\
    \textbf{Relations:} http://dbpedia.org/ontology/birthPlace}
    \end{center}

\end{itemize}

\textbf{German:}
\begin{itemize}
    \item \begin{center}
    \textit{\textbf{Sentence:} danach kamen abgeleitete klassen hinzu ein strengeres typsystem und während stroustrup "c with classes'' ("c mit klassen'') entwickelte woraus später c++ wurde schrieb er auch cfront einen compiler der aus c with classes zunächst c-code als erzeugte}
    \\
    \textit{\textbf{Entities:} (c,c++)
    \\
    \textbf{Relations:} http://dbpedia.org/ontology/influenced}
    \end{center}
    \item \begin{center}
    \textit{\textbf{Sentence:} früher auch ur ist ein 96.1 km langer nebenfluss der sauer entlang der grenze von deutschland zu den westlichen nachbarstaaten belgien und luxemburg }
    \\
    \textit{\textbf{Entities:} (sauer, deutschland)
    \\
    \textbf{Relations:} http://dbpedia.org/ontology/locatedInArea}
    \end{center}
\end{itemize}

\textbf{French:}
\begin{itemize}
    \item \begin{center}
    \textit{\textbf{Sentence:} à la mort de boleslas v le pudique duc princeps de pologne la guerre civile en mazovie empêche conrad de revendiquer le trône de cracovie}
    \\
    \textit{\textbf{Entities:} (boleslas v le pudique, cracovie)
    \\
    \textbf{Relations:} http://dbpedia.org/ontology/deathPlace}
    \end{center}
    \item \begin{center}
    \textit{\textbf{Sentence:} les entreprises masson masson est le dirigeant effectif des trois entreprises du groupe cette situation se reflète désormais dans l actionnariat et les raisons sociales des sociétés qui deviennent joseph masson sons and company (montréal) masson langevin sons and company (québec) masson sons and company (glasgow) cette dernière société basée en écosse a surtout vocation de gérer les achats }
    \\
    \textit{\textbf{Entities:} (joseph masson, québec)
    \\
    \textbf{Relations:} http://dbpedia.org/ontology/birthPlace}
    \end{center}
\end{itemize}

\textbf{Spanish:}
\begin{itemize}
    \item \begin{center}
    \textit{\textbf{Sentence:} en 2003 apareció en anything else película de woody allen junto a christina ricci y jason biggs además actuó en la película para televisión l}
    \\
    \textit{\textbf{Entities:} (anything else, jason biggs)
    \\
    \textbf{Relations:} http://dbpedia.org/ontology/starring}
    \end{center}
    \item \begin{center}
    \textit{\textbf{Sentence:} es una comuna y población de francia en la región de borgoña departamento de yonne en el distrito de sens y cantón de sens-ouest }
    \\
    \textit{\textbf{Entities:} (sens, yonne)
    \\
    \textbf{Relations:} http://dbpedia.org/ontology/department}
    \end{center}
\end{itemize}

\begin{table}
\centering
\begin{tabular}{lcccc}
\hline
\textbf{Language} & \textbf{\#sentences} & \textbf{\# bags} & \textbf{\# non-NA bags} & \textbf{Average bag-size}\\
\hline
\verb|English| & {532499} & {216806} & {66932} & {4.50}\\
\verb|French| & {409087} & {226418} & {83951} & {2.88} \\
\verb|Spanish| & {456418} & {229512} & {80706} & {2.88}\\ 
\verb|German| & {438315} & {194942} & {45908} & {3.48}\\ 
\hline
\end{tabular}
\caption{Key statistics for DiS-ReX}
\label{tab:4}
\end{table}

\begin{table}
\centering
\begin{tabular}{lc}
\hline
\textbf{Number of languages} & \textbf{Number of Bags}\\
\hline
\verb 2 & {59709}\\
\verb 3 & {9494}\\
\verb 4 & {1488}\\ 
\hline
\end{tabular}
\caption{Number of bags common across 2,3 and all languages}
\label{tab:5}
\end{table}

\begin{table*}
\centering
\begin{tabular}{|*{9}{c|} }
\hline 
Relation Label & English & French & German & Spanish & All languages \\
\hline 
NA & 149874 & 142467 & 149034 & 148806 & 590181 \\
 \hline 
isPartOf & 2548 & 645 & 465 & 490 & 4148 \\ 
 \hline 
state & 1882 & 1762 & 3537 & 429 & 7610 \\ 
 \hline 
largestCity & 265 & 342 & 199 & 393 & 1199 \\ 
 \hline 
birthPlace & 7861 & 9532 & 3341 & 9484 & 30218 \\ 
 \hline 
deathPlace & 4377 & 5629 & 277 & 4709 & 14992 \\ 
 \hline 
nationality & 2205 & 4413 & 143 & 2265 & 9026 \\ 
 \hline 
country & 10024 & 9618 & 3065 & 9808 & 32515 \\ 
 \hline 
capital & 544 & 651 & 397 & 891 & 2483 \\ 
 \hline 
city & 1415 & 4257 & 7930 & 1844 & 15446 \\ 
 \hline 
author & 1483 & 1224 & 94 & 460 & 3261 \\ 
 \hline 
previousWork & 348 & 696 & 305 & 1127 & 2476 \\ 
 \hline 
location & 5655 & 1300 & 1180 & 1685 & 9820 \\ 
 \hline 
riverMouth & 464 & 880 & 3303 & 154 & 4801 \\ 
 \hline 
locatedInArea & 1324 & 785 & 5715 & 608 & 8432 \\ 
 \hline 
hometown & 1689 & 435 & 163 & 4474 & 6761 \\ 
 \hline 
successor & 1574 & 2959 & 74 & 1618 & 6225 \\ 
 \hline 
influenced & 820 & 453 & 61 & 188 & 1522 \\ 
 \hline 
headquarter & 1122 & 922 & 460 & 1895 & 4399 \\ 
 \hline 
province & 225 & 1121 & 1272 & 2405 & 5023 \\ 
 \hline 
associatedBand & 3669 & 384 & 107 & 2555 & 6715 \\ 
 \hline 
subsequentWork & 390 & 760 & 344 & 1248 & 2742 \\ 
 \hline 
locationCountry & 925 & 799 & 2237 & 361 & 4322 \\ 
 \hline 
bandMember & 1327 & 1909 & 300 & 3092 & 6628 \\ 
 \hline 
director & 1258 & 3003 & 1592 & 2089 & 7942 \\ 
 \hline 
team & 1329 & 564 & 461 & 634 & 2988 \\ 
 \hline 
artist & 1188 & 3891 & 1241 & 2670 & 8990 \\ 
 \hline 
related & 1439 & 375 & 117 & 6262 & 8193 \\ 
 \hline 
producer & 1381 & 2848 & 1401 & 3044 & 8674 \\ 
 \hline 
predecessor & 475 & 2814 & 81 & 273 & 3643 \\ 
 \hline 
leaderName & 353 & 236 & 270 & 223 & 1082 \\ 
 \hline 
formerBandMember & 960 & 1153 & 174 & 1345 & 3632 \\ 
 \hline 
recordLabel & 791 & 881 & 199 & 2107 & 3978 \\ 
 \hline 
region & 1529 & 3673 & 1907 & 2249 & 9358 \\ 
 \hline 
influencedBy & 954 & 533 & 86 & 291 & 1864 \\ 
 \hline 
starring & 3040 & 7018 & 3087 & 4179 & 17324 \\ 
 \hline 
department & 99 & 5486 & 323 & 3157 & 9065 \\ 
 \hline 
All relations & 216806 & 226418 & 194942 & 229512 & 876743 \\ 
 \hline 
\end{tabular}
\caption{Comprehensive bag-wise statistics of the dataset}
\label{tab:6}
\end{table*}

\begin{table*}
\centering
\begin{tabular}{|*{9}{c|} }
\hline 
Relation Label & English & French & German & Spanish & All languages \\ 
\hline 
NA & 231271 & 167509 & 278360 & 224156 & 901296\\ 
\hline 
isPartOf & 16085 & 2794 & 2566 & 1880 & 23325\\ 
\hline 
state & 11979 & 13135 & 13705 & 1405 & 40224\\ 
\hline 
largestCity & 18811 & 4163 & 8949 & 3136 & 35059\\ 
\hline 
birthPlace & 15738 & 16624 & 4376 & 14359 & 51097\\ 
\hline 
deathPlace & 11498 & 12208 & 539 & 8888 & 33133\\ 
\hline 
nationality & 5848 & 9560 & 219 & 4330 & 19957\\ 
\hline 
country & 88787 & 43911 & 13148 & 64660 & 210506\\ 
\hline 
capital & 19887 & 4713 & 17227 & 5318 & 47145\\ 
\hline 
city & 4490 & 11156 & 23631 & 3740 & 43017\\ 
\hline 
author & 3387 & 4121 & 335 & 1417 & 9260\\ 
\hline 
previousWork & 6507 & 1276 & 450 & 2318 & 10551\\ 
\hline 
location & 15538 & 4757 & 4656 & 6014 & 30965\\ 
\hline 
riverMouth & 1172 & 2442 & 12467 & 420 & 16501\\ 
\hline 
locatedInArea & 4320 & 4152 & 18890 & 1904 & 29266\\ 
\hline 
hometown & 7648 & 796 & 1067 & 8971 & 18482\\ 
\hline 
successor & 4700 & 6963 & 128 & 3118 & 14909\\ 
\hline 
influenced & 2416 & 1147 & 635 & 394 & 4592\\ 
\hline 
headquarter & 5419 & 2399 & 2030 & 5736 & 15584\\ 
\hline 
province & 1082 & 2472 & 2710 & 11672 & 17936\\ 
\hline 
associatedBand & 7390 & 713 & 136 & 8437 & 16676\\ 
\hline 
subsequentWork & 6541 & 1318 & 517 & 2526 & 10902\\ 
\hline 
locationCountry & 3204 & 2836 & 8226 & 1229 & 15495\\ 
\hline 
bandMember & 3592 & 5910 & 475 & 8763 & 18740\\ 
\hline 
director & 2005 & 7811 & 2970 & 3961 & 16747\\ 
\hline 
team & 1830 & 814 & 694 & 1396 & 4734\\ 
\hline 
artist & 2893 & 9591 & 3156 & 6472 & 22112\\ 
\hline 
related & 4526 & 928 & 171 & 17432 & 23057\\ 
\hline 
producer & 2459 & 6398 & 2647 & 6384 & 17888\\ 
\hline 
predecessor & 2592 & 7003 & 162 & 600 & 10357\\ 
\hline 
leaderName & 1549 & 1074 & 452 & 448 & 3523\\ 
\hline 
formerBandMember & 2975 & 3452 & 279 & 4091 & 10797\\ 
\hline 
recordLabel & 1320 & 1214 & 219 & 4149 & 6902\\ 
\hline 
region & 5836 & 11860 & 5901 & 4485 & 28082\\ 
\hline 
influencedBy & 2524 & 1482 & 913 & 536 & 5455\\ 
\hline 
starring & 4484 & 14578 & 4616 & 6676 & 30354\\ 
\hline 
department & 196 & 15807 & 693 & 4997 & 21693\\ 
\hline 
All relations & 532499 & 409087 & 438315 & 456418 & 1858012\\ 
\hline 
\end{tabular}
\caption{Comprehensive sentence-wise statistics of the dataset}
\label{tab:7}
\end{table*}
\end{document}